\documentclass{article}
\usepackage{ijcai22}

\usepackage{times}

\usepackage{soul}
\usepackage{url}
\usepackage[hidelinks]{hyperref}
\usepackage[utf8]{inputenc}
\usepackage[small]{caption}
\usepackage{graphicx}
\usepackage{amsmath}
\usepackage{booktabs}

\usepackage{algorithm}
\usepackage{algorithmic}

\long\def\comment#1{}

\title{Federated Unlearning with Knowledge Distillation}

\author{
Chen Wu$^1$\and
Sencun Zhu$^1$\and
Prasenjit Mitra$^1$\\
\affiliations
$^1$The Pennsylvania State University\\
\emails
\{cvw5218, sxz16, pum10\}@psu.edu
}

\begin{document}

\maketitle

\begin{abstract}
Federated Learning (FL) is designed to protect the data privacy of each client during the training process by transmitting only models instead of the original data.
However, the trained model may memorize certain information about the training data. 
With the recent legislation on \textit{right to be forgotten}, it is crucially essential for the FL model to possess the ability to forget what it has learned from each client. 
We propose a novel federated unlearning method to eliminate a client's contribution by subtracting the accumulated historical updates from the model and leveraging the knowledge distillation method to restore the model's performance without using any data from the clients.
This method does not have any restrictions on the type of neural networks and does not rely on clients' participation, so it is practical and efficient in the FL system. We further introduce backdoor attacks in the training process to help evaluate the unlearning effect.
Experiments on three canonical datasets demonstrate the effectiveness and efficiency of our method. 
\end{abstract}

\section{Introduction}
The success of deep learning models relies on large-scale datasets.
Traditional training methods require collecting the training data and centralizing the data in one place. 
In many cases, due to personal wills or privacy laws, e.g., medical records from a hospital \cite{DBLP:journals/corr/LiuGNDKBVTNCHPS17}, the data cannot be moved from the clients. 
To build a global machine learning model from data obtained from multiple sources that cannot send the data to the server, we need to push the computation iteratively to the edge.
To solve this problem, researchers proposed a training method called Federated Learning (FL) \cite{DBLP:conf/aistats/McMahanMRHA17} to enable collaborative model training under such constraints.

However, 
certain real-world situations may demand that a client's contribution be removed from the global model that has been trained. 
From privacy perspective, recent legislation like the General Data Protection Regulation (GDPR) from the European Union \cite{voigt2017eu} and the California Consumer Privacy Act (CCPA) \cite{pardau2018california} from the United States both require the \textit{right to be forgotten} \cite{DBLP:conf/hotcloud/ShastriWC19}.
It grants users an unconditional right to request that their private data be removed from everywhere in the system within a reasonable time. 
The challenge is that the global machine learning model may still retain information about the client even though the original data has never been shared. 
The predictions made by the global model may leak the client information \cite{DBLP:conf/ndss/Salem0HBF019,DBLP:conf/cvpr/ZhangJP0LS20} and thus violate the legislation \textit{right to be forgotten}. 
Therefore there is a great need for approaches to remove a client's contribution from the trained global model.  

From the security perspective, removing a client's contribution is also essential when the client is intentionally malicious or unintentionally using outdated or low-quality data in the training process. 
Previous studies have shown that FL system can be compromised by data poisoning attacks from malicious clients \cite{DBLP:conf/icml/BhagojiCMC19,DBLP:conf/aistats/BagdasaryanVHES20,DBLP:conf/iclr/XieHCL20}.
Thus, the ability to completely remove the influence of such clients can significantly improve the security and reliability of the FL system.

A naive way to make the global model provably forget the contribution from a specific client is to retrain the model from scratch.
However, this can result in an enormous cost of time and energy.
Existing works on machine unlearning tasks mostly focus on centralized learning scenario with free access to the training dataset \cite{DBLP:conf/ccs/DuCLOS19,DBLP:conf/sp/BourtouleCCJTZL21} or does not work for complex models such as deep neural networks (DNNs) \cite{DBLP:conf/sp/CaoY15,DBLP:conf/nips/GinartGVZ19,DBLP:conf/aistats/IzzoSCZ21}. 
The only unlearning work in the FL scenario provides limited improvement over retraining from scratch since it requires clients to retrain the model to adjust their historical updates \cite{DBLP:conf/iwqos/LiuMYWL21}.
In federated learning, each learning iteration of the global model consists of an aggregation of gradient updates from participating clients. 
Reducing the number of iterations between the server and clients is crucially important since the communication costs a tremendous amount of time and energy, especially for DNNs. 
Also, to improve the efficiency of FL, it is always better to put more computation on the server's side rather than on the client's side because the server usually has more computation power than clients. 
So, designing an efficient federated unlearning method should also follow these two rules. 

In this paper, we define the federated unlearning problem as removing a designated client's contribution thoroughly and efficiently from the global model after the federated training process. 
However, evaluating the effectiveness of the unlearning task is not easy since there are too many stochastic processes in the FL system -- even retraining the model may produce different results each time. 
To intuitively evaluate the unlearning effect, we introduce backdoor attack \cite{DBLP:journals/corr/abs-1708-06733,DBLP:conf/aistats/BagdasaryanVHES20} in the client's updates. 
As one of the most powerful attacks to the FL system, the backdoor attack does not influence the global model's performance on regular input and only distorts the predictions when triggered by specific inputs with the backdoor pattern. 
Such a property makes it a perfect evaluation method to measure the effect of unlearning. 
A successful unlearning global model should perform well on the evaluation dataset but reduce backdoor attack's success rate when triggered by backdoored inputs. 

To tackle this challenge, we propose a novel federated unlearning method that can eliminate the attacker's influence and vastly reduce the unlearning cost in the FL system.
The idea is to erase the historical parameter updates from the attacker and recover the damage through the knowledge distillation method. 
Specifically, we use the old global model as a teacher to train the unlearning model. 
This approach has many advantages.
Firstly, the knowledge distillation training is operated entirely on the server's side without requiring a labeled dataset, so there will be no client-side time and energy costs and no network transmission.
Secondly, the backdoor features will not transfer from the teacher model to unlearning model since those features will not activate without emerging of backdoor patterns \cite{DBLP:journals/corr/abs-1708-06733}. 
Lastly, distillation prevents the model from fitting too tightly to the data and contributes to a better generalization around training points \cite{DBLP:conf/sp/PapernotM0JS16}, so it can help improve the robustness of the model and further improve the model's performance with post-training afterward.
In the end, empirical studies on three datasets using different DNN architectures have all proved the effectiveness of our federated unlearning method.

\section{Related Work}
The work of Cao and Yang introduces the term ``machine unlearning'' \cite{DBLP:conf/sp/CaoY15}. They presented an unlearning algorithm by transforming the learning algorithm into a summation form. To forget a training data sample, they update a small portion of the summations, and thus it is asymptotically faster than retraining from scratch. However, this algorithm only works for traditional machine learning methods that can be transformed into a summation form. 
Ginart et al. formalized the problem and notion of efficient data deletion in machine learning and propose two efficient deletion solutions for the k-means clustering algorithm \cite{DBLP:conf/nips/GinartGVZ19}. 
Bourtoule et al. introduced SISA training to reduce the computational overhead associated with unlearning \cite{DBLP:conf/sp/BourtouleCCJTZL21}. This framework uses data sharding and slicing to strategically limit the influence of a data point in the training procedure. 
Izzo et al. proposed an approximate data deletion method called projective residual update (PRU) for linear and logistic models \cite{DBLP:conf/aistats/IzzoSCZ21}. The computational cost is linear in the feature dimension and independent of training data size. 
Du et al. proposed an unlearning framework to correct the model when a false negative or false positive case is labeled under lifelong anomaly detection task \cite{DBLP:conf/ccs/DuCLOS19}.
Liu et al. studied the unlearning problem in Federated Learning scenario \cite{DBLP:conf/iwqos/LiuMYWL21}.
Their method is to adjust the historical parameter updates of federated clients through the retraining process in FL and reconstruct the unlearning model.
It relies on the clients' participation (with their historical dataset) and needs extra rounds of communication between clients and the server.


\section{Problem Definition}
As required by privacy regulations such as the GDPR and the CCPA, individuals have an unconditional right to request the removal of their private data from the system within a reasonable time.
This requirement challenges the existing FL system with the ability to erase the contribution of clients from the training process. 
We first introduce the Federated Learning (FL) system and define the objective of unlearning.
Then, we discuss the challenges of unlearning in the FL system and show the 
limitations of existing unlearning methods. 
In the end, we introduce a novel method to evaluate the unlearning effect using backdoor attacks in the FL system.

\subsection{Federated Learning and Unlearning}
In a general perspective, the problem includes two entities: the server $S$ and a group of clients $C$ participated in the FL.
The server $S$ relies on the group of clients $C$ to help train a global model $M$ to be used by all the clients.
According to the FL definition~\cite{DBLP:conf/aistats/McMahanMRHA17}, training data is on the clients' side and cannot be shared with others. 
Each client will train the global model $M$ on its local dataset and send its model updates to the server.
The server will aggregate model updates from clients and generate an updated global model for the next round of training.
Suppose a client $i \in C$ needs to revoke its previous contribution/updates to the global model $M$. In that case, the server should be able to update the current global model $M$ to an unlearned model $M^{C \setminus \{i\}}$, where $M^{C \setminus \{i\}}$ is a model that could be trained if client $i$ was never included in the training group $C$.
 
Assume that there are a total of $N$ clients to participate in the FL training at round $t$, and each client trains the global model $M_{t}$ with its private dataset and updates parameter changes to the server. 
The new global model $ M_{t+1}$ is calculated by an averaged aggregation (\textit{FedAvg} \cite{DBLP:conf/aistats/McMahanMRHA17}) of the weight updates from these clients.
\vspace{-1ex}
\begin{equation} \label{eq:fedavg_update_rule}
    M_{t+1} = M_{t} + \frac{1}{N} \sum_{i=1}^{N} \Delta M_{t}^{i}
    \vspace{-1ex}
\end{equation}
where $\Delta M_{t}^{i}$ is the parameter update contributed by client $i$ based on the model $M_{t}$. 
The server keeps running the training process and updates the global model from $M_{1}$ to $M_{F}$ when the termination criterion (e.g. test accuracy) has been satisfied at round $F-1$. 

The federated unlearning task is defined as to completely remove the influence of all the updates $\Delta M_{t}^{i}$ from a target client $i$, with $t \in [1, F-1]$ from the model $M_{F}$.
In other words, it removes the contribution of any client $i$ to the final global model $M_{F}$ and creates a new model $M_{F}^{C \setminus \{i\}}$ as if client $i$ has never participated in the training process.

\subsection{Challenges in Federated Unlearning}
\label{challenges}

\noindent \textbf{Incremental Learning Process}
The update of the model is an incremental process where any update depends on all previous updates. 
For example, if $\Delta M_{T}^{k}$, the update from client $k$, was removed from the global model aggregation process at round $T$, the global model $M_{T+1}$ would change (based on Equ. (1)) to a new model $M_{T+1}'$. 
Because each client $i$ originally calculated its local update $\Delta M_{T+1}^{i}$ based on the global model $M_{T+1}$, all its subsequent updates $\Delta M_{t}^{i}$ ($t \geq T+1$) became invalid after the removal of client $k$.
In this case, if $T=1$, i.e., the first round of the FL process, we may need to retrain the model from scratch with all clients participating and recalculating the corresponding changes in the later updates. 

\noindent \textbf{Stochastic Training Process}
There is much more randomness exists in the FL training process than in a centralized training process.
Firstly, the clients who participated in each round of training are randomly selected. 
Besides, there is a lot of randomness in the local training process of each client, like the randomly sampled small batches of data and the ordering of the batches. 
A tiny perturbation in any of the above processes will cause a butterfly effect in the following training process.
This makes the FL process non-deterministic and hard to control. 
Even retraining the model from scratch may lead the global model to converge to different local minima each time. 

\noindent \textbf{Limited Access to Dataset}
Another obstacle in FL is that server does not have access to the training data possessed by clients.
Thus, the previous unlearning techniques that rely on dataset splitting cannot be applied to the FL scenario \cite{DBLP:conf/sp/BourtouleCCJTZL21}.
A more practical problem is that the edge devices of clients may only have limited storage space, and they may delete the data anytime after the training process. 
It will make even the most naive way of retraining the model from scratch impractical since the client may not have the same data as that used in the training time. 
Similar methods that ask clients to calibrate the historical updates by retraining the model again may also only work in theory \cite{DBLP:conf/sp/BourtouleCCJTZL21}.

\subsection{Evaluation of Federated Unlearning}
With the non-deterministic property of FL, it is natural to ask the question of how to evaluate the unlearning model.
Is there a way to measure the unlearning effect in FL system?
Previous unlearning works on DNNs \cite{DBLP:conf/sp/BourtouleCCJTZL21,DBLP:conf/sp/BourtouleCCJTZL21} compare the test accuracy between unlearning model and retrained model may not be convincing enough in FL systems. 
Therefore, we introduce the backdoor attacks \cite{DBLP:journals/corr/abs-1708-06733,DBLP:conf/aistats/BagdasaryanVHES20} to help evaluate the effectiveness of unlearning methods. 
As one of the most powerful attacks to the FL system, the backdoor attack does not influence the global model's performance on regular input but only distorts the predictions when triggered by specific inputs with the backdoor pattern.
Such property makes it a perfect evaluation method to measure the effect of unlearning. 
A successful unlearning global model should perform reasonably well on the evaluation dataset but reduce backdoor attacks' success rate when triggered by backdoored inputs. 

\section{Unlearning Method}
To tackle the challenges outlined in Section~\ref{challenges}, we propose a novel federated unlearning method, as shown in Algorithm \ref{alg:algorithm}, that can eliminate the client's contribution and vastly reduce the unlearning cost in the FL system.
This method requires the server to keep the history of parameter updates from each contributing client and possess some extra outsourced unlabeled data. 
The key idea is to first erase the historical parameter updates from the target client and then recover the damage through the knowledge distillation method. 

\begin{algorithm}[tb]
\caption{Federated Unlearning}
\label{alg:algorithm}
\textbf{Input}: Global model $M_{F}$, Total number of clients $N$\\
\textbf{Input}: Historical updates $\Delta M_{t}^{A}$ of target client $A$ at round $t$\\
\textbf{Input}: Outsourced unlabelled dataset $X$ \\
\textbf{Parameter}: Distillation epoch $k$, Temperature $T$\\
\textbf{Output}: The unlearning model $M_{F}'$
\begin{algorithmic}[1] 
\STATE $M_{F}' \leftarrow M_{F} - \frac{1}{N} \sum_{t=1}^{F-1} \Delta M_{t}^{A}$
    \FOR {$epoch = 1, 2, \ldots, k$}
        \STATE $y_{teacher} \leftarrow M_{F}(X), T$
        \STATE $y_{student} \leftarrow M_{F}'(X), T$
        \STATE Calculate $loss_{distillation}$ of $y_{teacher}$ and $y_{student}$
        \STATE Back-propagate model $M_{F}'$
    \ENDFOR
\STATE \textbf{return} unlearning model $M_{F}'$
\end{algorithmic}
\end{algorithm}

\begin{figure*}
    \centering
    \includegraphics[width=0.95\textwidth]{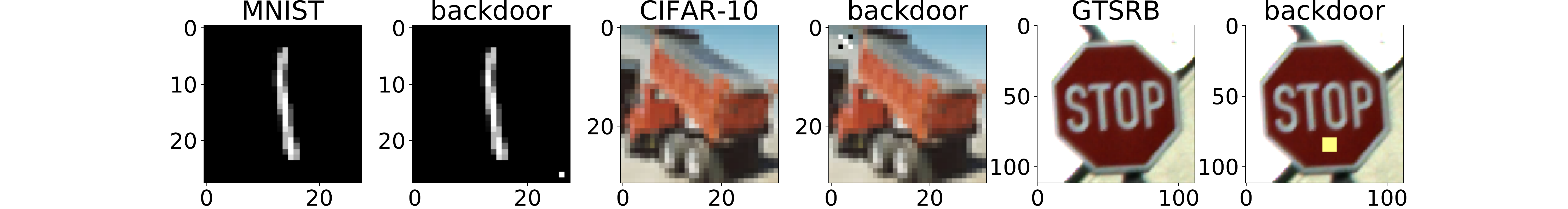}
    \caption{Samples of backdoor images targeting digit 1 in MNIST, class of truck in CIFAR-10, and the class of stop sign in GTSRB dataset.}
    \label{fig:backdoor_samples}
    \vspace{-2ex}
\end{figure*}

\subsection{Erase Historical Parameter Updates}
To completely remove the contribution of any client $i$ to the final global model $M_{F}$, we want to erase all the historical updates $\Delta M_{t}^{i}$ from this client as long as $t \in [1, F-1]$. 
As we have shown in Equation \ref{eq:fedavg_update_rule}, the update of the global model at round $t$ consists of averaged weight updates from participating clients. If we use $\Delta M_{t}$ to represent this update at round $t$, the finalized global model $M_{F}$ can be viewed as a composition of initial model weight $M_{1}$ and updates to the global model from round $1$ to round $F-1$. 
\vspace{-1ex}
\begin{equation} \label{eq:final_model_update_rule}
    M_{F} = M_{1} + \sum_{t=1}^{F-1} \Delta M_{t}
    \vspace{-1ex}
\end{equation}

For simplicity, we assume that each round, there are $N$ clients participating in the FL training, and client $N$ is the target client that wants to be unlearned from the global model. At this time, we can simplify the problem as to remove the contribution $\Delta M_{t}^{N}$ of the target client $N$ from the global model update $\Delta M_{t}$ at each round $t$. 
\vspace{-1ex}
\[
    \Delta M_{t} = \frac{1}{N} \sum_{i=1}^{N} \Delta M_{t}^{i} = \frac{1}{N} \sum_{i=1}^{N-1} \Delta M_{t}^{i} + \frac{1}{N} \Delta M_{t}^{N}
    \vspace{-1ex}
\]

There are two ways to calculate the new global model update $\Delta M_{t}'$ at round $t$. The first one is to assume that only $N-1$ clients were participating in the FL at round $t$. In this way, the new global model updates $\Delta M_{t}'$ at round $t$ becomes the following equation.
\vspace{-1ex}
\[
    \Delta M_{t}' = \frac{1}{N-1} \sum_{i=1}^{N-1} \Delta M_{t}^{i} = \frac{N}{N-1} \Delta M_{t} - \frac{1}{N-1} \Delta M_{t}^{N}
    \vspace{-1ex}
\]

However, we can not directly accumulate the new updates to reconstruct the unlearning model because of the incremental learning property of FL, as discussed in the previous section. Any update to the global model $M_{t}$ will result in a requirement of updates to all the model updates that happened afterward. Hence, we use $\epsilon_{t}$ to represent the necessary amendment (skew) to the global model at each round $t$. 
After combining the above equation with Equation \ref{eq:final_model_update_rule}, we can get the unlearning version of the final global model $M_{F}'$.
\vspace{-1ex}
\[
    M_{F}' = M_{1} + \frac{N}{N-1} \sum_{t=1}^{F-1} \Delta M_{t} - \frac{1}{N-1} \sum_{t=1}^{F-1} \Delta M_{t}^{N} + \sum_{t=1}^{F-1} \epsilon_{t}
    \vspace{-1ex}
\]
where $\epsilon_{t}$ is the necessary correction to amend the skew produced by change of the model at previous rounds. 
Because of this characteristic of the incremental learning process in FL, the skew $\epsilon_{t}$ will increase with more training rounds after updates to the global model. Thus, the above unlearning rule has a shortcoming that when the target client $N$ makes little contribution to the model at round $t$ (e.g., $\Delta M_{t}^{N} \approx 0$), the global model update $\Delta M_{t}$ will still change a lot by multiplying itself with a factor of $\frac{N}{N-1}$. This will bring more skew $\epsilon_{t}$ to the global model in the following rounds. 

To mitigate this problem, we propose to use a lazy learning strategy to eliminate the influence of target client $N$. 
Specifically, we assume client $N$ still participated in the training process but set his updates $\Delta M_{t}^{N} = 0$ for all rounds $t \in [1, F-1]$. 
The unlearning of the global model update can be simplified as follows.
\vspace{-1ex}
\[
    \Delta M_{t}' = \frac{1}{N} \sum_{i=1}^{N-1} \Delta M_{t}^{i} = \Delta M_{t} - \frac{1}{N} \Delta M_{t}^{N}
    \vspace{-1ex}
\]
A combination of the above formula with Equation \ref{eq:final_model_update_rule} gives us the unlearning result of the final global model $M_{F}'$.
\vspace{-1ex}
\begin{flalign} \label{eq:unlearning_model_update_rule}
    M_{F}' &= M_{1} + \sum_{t=1}^{F-1} \Delta M_{t}' + \sum_{t=1}^{F-1} \epsilon_{t} &&\nonumber\\
    &= M_{1} + \sum_{t=1}^{F-1} \Delta M_{t} - \frac{1}{N} \sum_{t=1}^{F-1} \Delta M_{t}^{N} + \sum_{t=1}^{F-1} \epsilon_{t} &&\nonumber\\
    &= M_{F} - \frac{1}{N} \sum_{t=1}^{F-1} \Delta M_{t}^{N} + \sum_{t=1}^{F-1} \epsilon_{t}
    \vspace{-2ex}
\end{flalign}

Now the unlearning model update rule becomes surprisingly straightforward and easy to understand. We just need to subtract all the historical averaged updates from target client $N$ from the final global model $M_{F}$. Then, we remedy the skew $\epsilon_{t}$ caused by this process because of the incremental learning characteristic of the FL. 

\subsection{Remedy with Knowledge Distillation}
There is no existing method to calculate the skew $\epsilon_{t}$ without retraining the updated model on the original dataset again. 
However, as we discussed before, one of the challenges in FL is that we cannot rely on the clients to hold the dataset forever and prepare for the unlearning purpose. 
How to remedy this skew without relying on clients' training data or the participation of clients becomes crucially important. 

To tackle this problem, we propose to leverage the knowledge distillation method to train the unlearning model using the original global model. 
The distillation technique is first motivated by the goal of reducing the size of DNN architectures or ensembles of models \cite{DBLP:journals/corr/HintonVD15}. 
It uses the prediction results of class probabilities produced by an ensemble of models or a complex DNN to train another DNN of a reduced number of parameters without much loss of accuracy.
Later on, Papernot et al. proposed to use this method as a defensive mechanism to reduce the effectiveness of adversarial samples on DNNs \cite{papernot2016distillation}.
They showed that the defensive distillation can improve the generalizability and robustness of the trained DNNs. 
The intuition is based on the fact that the knowledge acquired by DNNs during the training process is not only encoded in the weight parameters but also can be reflected from the class probability prediction output of the model. 
Distillation training uses these soft probabilities instead of hard ground truth labels to provide additional information about each class and the prediction logic to the training process. 
For example, given an image of ``7'' from the MNIST dataset, the distillation training process can train the new model with more information (except from the ground truth label that the image belongs to class ``7''), such as this image is close to ``1'' and ``9'' but a lot different from ``3'' and ``8''.

\begin{figure}
    \centering
    \includegraphics[width=0.45\textwidth]{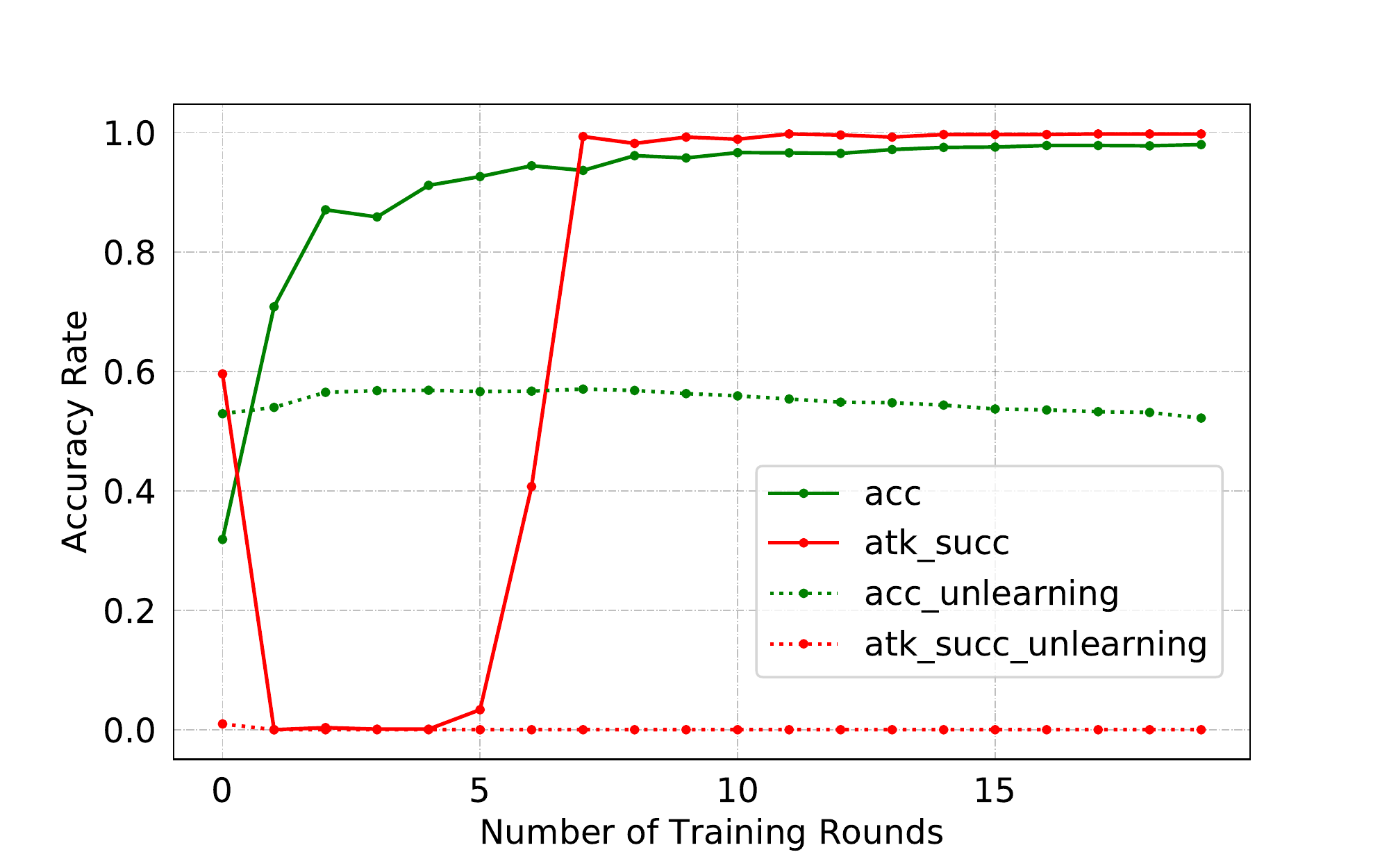}
    \caption{The training process and the corresponding subtraction unlearning process with MNIST dataset. The solid lines stand for the test accuracy and backdoor attack success rate in the training process. The dotted lines represent the test accuracy and backdoor attack success rate of the unlearning model after merely subtracting the target historical parameter updates.}
    \label{fig:unlearning_MNIST_step1}
    \vspace{-2ex}
\end{figure}

To perform knowledge distillation in the unlearning problem, we treat the original global model as the teacher model and the skewed unlearning model as the student model. 
Then, the server can use any unlabeled data 
to train the unlearning model and remedy the skew $\epsilon_{t}$ caused by the previous subtraction process. 
Specifically, the original global model produces class prediction probabilities through a ``softmax'' output layer that converts the logit, $z_{i}$, computed for each class into a probability, $q_{i}$, by comparing $z_{i}$ with the other logits. 
\vspace{-1ex}
\begin{equation}
    q_{i} = \frac{\exp (z_{i} / T)}{\sum_{j} \exp (z_{j} / T)}
    \vspace{-1ex}
\end{equation}
where $T$ is a parameter named \textit{temperature} and shared across the softmax layer. 
The value of $T$ is normally set to 1 for traditional ML training and predictions. 
A higher temperature $T$ makes the DNN produce a softer probability distribution over classes.
In other words, the probability output will be forced to produce relatively large values for each class, and logits $z_{i}$ become negligible compared to temperature $T$. 
An example is that the output probability for each class $i$ converge to $1 / Z$ (assume there are $Z$ classes for prediction) as $T \rightarrow \infty$. 
In summary, higher temperature $T$ produces probability distribution more ambiguously while lower temperature $T$ produces probability distribution more discretely. 

We use this soft class prediction probability produced by the original global model $M_{F}$ to label the dataset.
The skewed unlearning model is then trained with this dataset with soft labels (with high temperature $T$).
On the other hand, if the server possesses a labeled dataset, we can also leverage a combination of hard labels (ground truth with temperature $T = 1$) and soft labels produced by the global model.  
As suggested by Hinton et al., using a weighted average of these two objective functions with a considerably lower weight on the objective function of the hard labels can produce the best results \cite{DBLP:journals/corr/HintonVD15}.
The temperature will be set back to $1$ after distillation training, so the unlearning model $M_{F}'$ can produce more discrete class prediction probabilities during test time.

\begin{figure}
    \centering
    \includegraphics[width=0.45\textwidth]{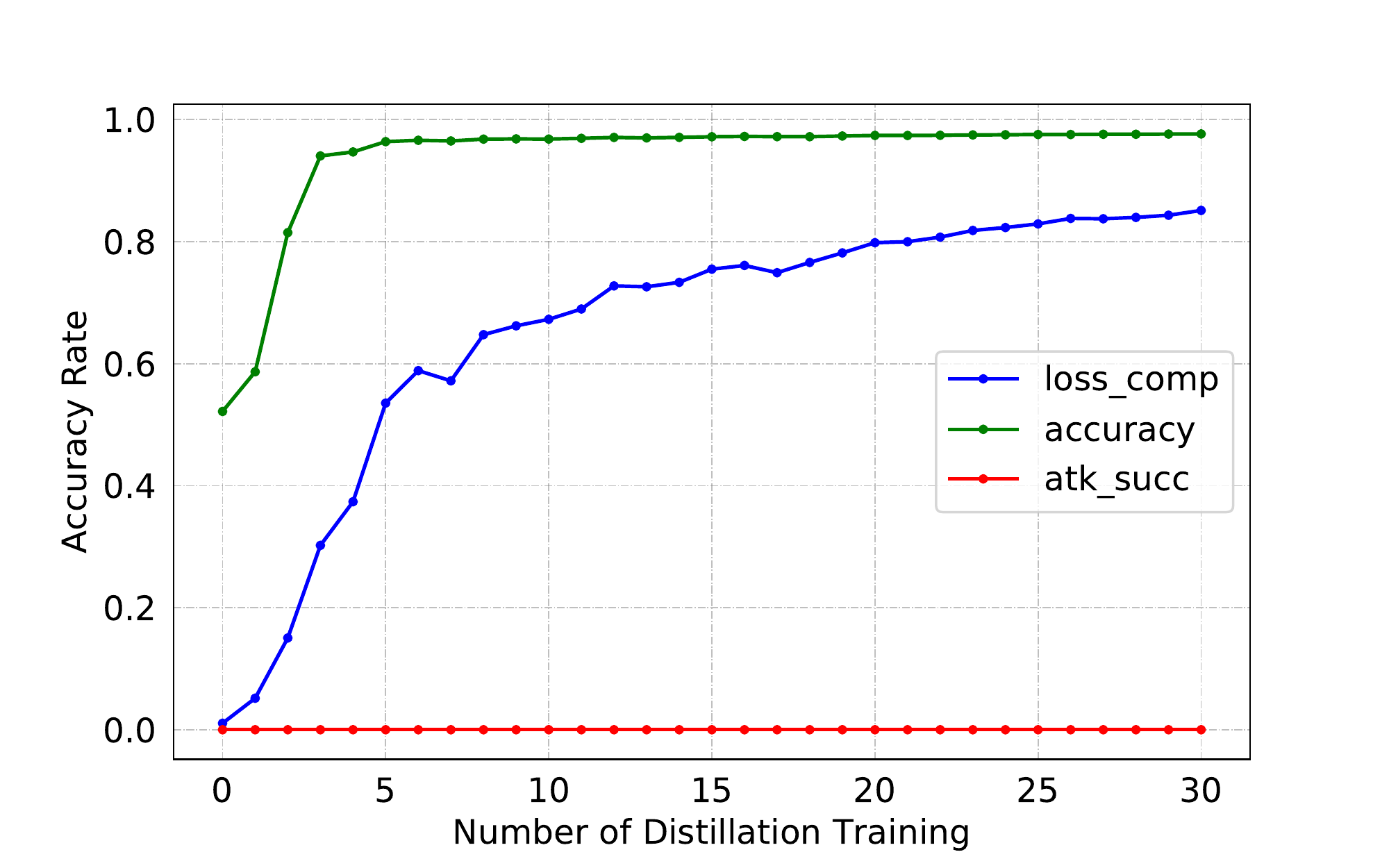}
    \caption{The performance of the unlearning model under knowledge distillation training process with MNIST dataset. The blue line stands for the change of losses which is calculated using the loss of the original model before subtracting historical parameter updates from the target client divided by the loss of the unlearning model. It means the model is better recovered with a value closer to 1.}
    \label{fig:unlearning_MNIST_step2}
    \vspace{-2ex}
\end{figure}

\section{Experiments}
We empirically evaluate the proposed unlearning method using different model architectures with three datasets.
The central results show that our unlearning strategies can effectively remove the contribution of the target client (attacker) from the global model.
Moreover, the damage to the model can be quickly recovered through the distillation training process purely on the server's side.

\begin{table*}[htb]
\caption{Validation Results of Unlearning Method under Different Stage}
\centering
\begin{tabular}{|c|c|c|c|c|c|c|c|c|c|c|}
    \hline
    \multicolumn{1}{|c|}{} & \multicolumn{2}{|c|}{\textbf{Training}} & \multicolumn{2}{|c|}{\textbf{UL-Subtraction}
    } & \multicolumn{2}{|c|}{\textbf{UL-Distillation}} & \multicolumn{2}{|c|}{\textbf{Post-Training}} & \multicolumn{2}{|c|}{\textbf{Re-Training}} \\ 
    \hline
    \hline
    \textbf{Dataset} & test acc & atk acc & test acc & atk acc & test acc & atk acc & test acc & atk acc & test acc & atk acc\\
    \hline
    MNIST & 98.0 & 99.7 & 52.2 & 0 & 97.7 & 0 & 98.5 & 0 & 98.2 & 0 \\
    CIFAR-10 & 80.8 & 99.4 & 10.0 & 0 & 78.8 & 6.4 & 81.4 & 7.3 & 79.5 & 7.0 \\
    GTSRB & 93.0 & 100 & 3.9 & 0 & 92.1 & 0 & 94.0 & 0 & 92.7 & 0 \\
    \hline
\end{tabular}
\label{table:integrated_results} 
\vspace{-1ex}
\end{table*}

\subsection{Overview of the Experimental Setup}

\subsubsection{Dataset Description}
We use the following three canonical ML datasets in our experiments: the MNIST \cite{lecun1998gradient}, CIFAR-10 \cite{krizhevsky2009learning}, and the German Traffic Sign Recognition Benchmark (GTSRB) dataset \cite{Stallkamp-IJCNN-2011}. 
In the MNIST experiment, we have 10 clients participating in the FL process and one of them is the target client (attacker).
In the CIFAR-10 experiment, we perform a normalization step for each image before training and testing processes. 
There are also 10 clients and one of them is the target client.
In the GTSRB experiment, we first crop each image to make its height and width into 112 pixels and then perform a normalization step for each image before processing.
There are only 5 clients in the FL process, and one of them is the attacker.
We also sample some data (excluding the clients' data) for the distillation training process.
In the following experiments, we do not use the actual label of these data (treat them as unlabeled data), and the size is the same as that of a client's dataset.



\subsubsection{Model Architectures}
In our FL settings, the clients and the server share the same model for each dataset. 
We use three different models for different classification tasks.
Specifically, the model for MNIST is consisted of 2 convolutional layers, 2 max pool layers, and 2 fully connected layers in the end for prediction output.
As for CIFAR-10, we use the well-known VGG11 network \cite{DBLP:journals/corr/SimonyanZ14a}, which consists of 8 convolutional layers, 5 max pool layers, followed by one fully connected layer to produce probability prediction output. 
On the GTSRB dataset, we use another famous AlexNet \cite{DBLP:conf/nips/KrizhevskySH12}, which is composed of 5 convolutional layers, 3 max pool layers, and 3 fully connected layers for output.

\subsubsection{Unlearning Target with Backdoor Attack}
We use backdoor attacks in the target client's updates to the global model as described before, so that we can intuitively investigate the unlearning effect based on the attack success rate of the unlearned global model.
The backdoor attack is triggered by backdoor patterns in the input image. 
The target client changes some pixels in the benign inputs to create a backdoor pattern, as shown in Figure \ref{fig:backdoor_samples}. 
Backdoor targets in the experiments include making digit ``1'' predicted as digit ``9'' in MNIST, ``truck'' predicted as ``car'' in CIFAR-10, and ``Stop Sign'' predicted as ``Speed limit Sign (120 km/h)'' in the GTSRB dataset.

\subsection{Unlearning Model Performance Evaluation}
In this section, we evaluate the performance of the unlearning model under different steps during unlearning. 
To begin with, we show, in Figure \ref{fig:unlearning_MNIST_step1}, the behavior of the model after directly erasing all the historical parameter updates from the target client (attacker).
Subtracting the parameter updates from the global model will create a non-negligible skew. 
As we observed from the figure, the unlearning model's accuracy has never exceeded 60\% and has a trend of slightly decreasing with more FL training rounds.
The advantage of this process is that it can thoroughly remove the influence of the target client (attacker) from the global model.
Compared with the original global model with an almost 100\% backdoor attack success rate, the unlearning model keeps the attack success rate 0\% all the time. 

Then, we can look at the following knowledge distillation process used to recover the skew caused by the subtraction process.
From Figure \ref{fig:unlearning_MNIST_step2}, we notice that the test accuracy of the model is quickly recovered within five epochs of distillation training. 
The loss on the test dataset keeps getting closer to the original global model while the backdoor attack success rate keeps as low as 0\% all the time. 
The attacker's influence on the global model does not transmit to the unlearning model after the knowledge distillation training process. 
The reason is that the distillation training method does not use any data from the target client (attacker), so the backdoor in the original model has not been triggered and thus will not be learned by the unlearning model.
This also demonstrates the privacy protection of a client in the case of right to be forgotten. because the influence of his contributions will also be removed from the new global model. 

In the end, Table \ref{table:integrated_results} reports the integrated experiment results with different datasets and model architectures.
The ``Training'' column reports the performance of global model $M_{F}$ on the evaluation dataset, including the test accuracy and backdoor attack success rate. 
The ``UL-Subtraction'' column reports the unlearning model's performance after merely subtracting the target client's historical parameter updates from the global model.
The ``UL-Distillation'' column reports the performance of unlearning model after the knowledge distillation process on the server's side.
The ``Post-Training'' column represents how much we can further improve the unlearning model if putting it back to the FL system and continue training without the participation of the target client (attacker). 
The ``Re-Training'' column is the widely used golden standard for the unlearning problem.
We retrain the model from scratch excluding the participation of the attacker.

From the results, we can conclude that subtracting the attacker's historical parameter updates eliminates his influence on the global model. In all cases, the attack success rate drops to zero.
The knowledge distillation training process can help remedy the skew caused by subtraction and recover the model's performance back to an acceptable standard. 
The test accuracy of the unlearning model after distillation is almost identical to that of the model retraining from scratch (with differences less than 1\%). 
In the post-training results, we find the distillation can even help improve the model's accuracy with follow-up training with clients.
What is more, we are delighted to observe that the distillation process does not pass the unlearning target's attributes from the original global model to the unlearning model. 
It supports our assumption that as long as we are not using the same dataset to activate the global model, the original global model's private information/attributes will not be learned by the unlearning model.
One may ask why the attack success rate is larger than zero for even the retraining model on the CIFAR-10 dataset.
It is caused by the prediction errors of the model (the test accuracy is only around 80\%).
In other words, the original images (without backdoor patterns) will be misclassified as the backdoor target label too. 
Thus, such errors will be automatically counted as backdoor success rates here.
Compared with the attack success rate of almost 100\% in the original global model, under this error rate (less than 7\%), we may consider it a successful unlearning of the target client's contribution.


\section{Conclusion and Future Work}
This paper analyzes the difficulties and challenges of unlearning tasks in a realistic federated learning scenario. 
We propose the first federated unlearning method that does not rely on the clients' participation and dataset.
Our method can fully distinguish the influence of any client's contribution to the global model by subtracting his historical parameter updates from the model.
Then, we use the knowledge distillation method to remedy the skew of the unlearning model caused by the subtraction.
We also introduce backdoor attacks to help evaluate the unlearning effect in FL.
Empirical studies from our experiments on three canonical datasets have demonstrated the effectiveness of our unlearning method. 
Since our method is purely conducted on the server's side, the running time and energy cost efficiency ultimately defeat other existing methods that require extra communication between clients and server. 
We envision our work as a start to study the unlearning problem in the FL system from another angle. 
There are multiple interesting topics we can study in the future, such as combining the distillation in the training process and distillation without any dataset.

\clearpage
\bibliographystyle{named}
\bibliography{main}

\end{document}